\documentclass{article}

\pdfoutput=1

\usepackage{arxiv}
\usepackage{times}  
\usepackage{helvet} 
\usepackage{courier}  
\usepackage[hyphens]{url}  
\usepackage{graphicx} 
\usepackage{graphicx}  
\usepackage{comment}
\usepackage{amsfonts}
\usepackage{mathtools}
\usepackage{amsthm}
\usepackage{times}
\usepackage{graphicx} 
\usepackage{subfigure} 
\usepackage{algorithm}

\usepackage{algorithmic}

\usepackage{algorithm}
\usepackage{algorithmic}
\usepackage{amsmath}
\newcommand{\argmin}{\operatornamewithlimits{argmin}}
\newcommand{\argmax}{\operatornamewithlimits{argmax}}
\newtheorem{theorem}{Theorem}
\newtheorem{lemma}[theorem]{Lemma}

\newcommand{\bigO}{\mathcal{O}}

\title{Adversarial Sleeping Bandit Problems with Multiple Plays: Algorithm and Ranking Application}

\author{
Jianjun Yuan\\
  Expedia Group \\
  1111 Expedia Group Wy W, Seattle, WA, US, 98119 \\
  \texttt{yuanx270@umn.edu} \\
  \And
  Wei Lee Woon \\
  Expedia Group \\
  1111 Expedia Group Wy W, Seattle, WA, US, 98119 \\
  \texttt{wwoon@expediagroup.com} \\
  \And
  Ludovik Coba \\
  Expedia Group \\
  407 St John St, London, UK
}

\begin{document}
\maketitle

\begin{abstract}
  This paper presents an efficient algorithm to solve the sleeping bandit with multiple plays problem in the context of an online recommendation system. The problem involves bounded, adversarial loss and unknown i.i.d. distributions for arm availability. The proposed algorithm extends the sleeping bandit algorithm for single arm selection and is guaranteed to achieve theoretical performance with regret upper bounded by $\bigO(kN^2\sqrt{T\log T})$, where $k$ is the number of arms selected per time step, $N$ is the total number of arms, and $T$ is the time horizon.
\end{abstract}

\section{Introduction}
Recommendation systems utilize machine learning (ML) algorithms and large data-sets to suggest personalized content or products to users \cite{Ricci2022}.
These systems analyze patterns in user data, such as past purchases, browsing history, and ratings, to generate recommendations that are tailored to each user's preferences and interests. 
Recommendation systems have become increasingly popular in recent years and are used in a variety of use cases and industries, including online shopping in e-commerce~\cite{Smith2017},
online video browsing in entertainment~\cite{amatriain2015recommender},
and online advertising in social media~\cite{10.1145/3460231.3474618}. 

To understand and predict user behavior, ML algorithms are frequently
used as these can achieve high levels of accuracy by analyzing many different
signals and features. However, these algorithms can also be computationally
demanding, and in practice, they are only applicable to a few hundred
listings at a time. Depending on the application, thousands of
relevant candidates may need to be considered, resulting in
unacceptably high model latencies. One potential solution is to divide
the recommendation system into two main components
\cite{covington2016deep}: 1) a lightweight \emph{candidate generation}
stage, where a small subset of the most promising candidates, and 2)
\emph{ranking} this reduced subset such that the most relevant items
are shown at the top.



The focus of this paper is on the first phase (candidate generation)
listed above. ML techniques for candidate generation are usually
trained offline and updated in large batches. As such, they can be
slow to react to changes in user preferences, and may require frequent
model retraining \cite{rendle2008online}. 
Multi-arm bandit algorithms have been studied extensively in the online learning literature \cite{auer2000using,auer2002finite,auer2002nonstochastic,vermorel2005multi,cesa2006prediction} due to their success in solving the online selection problem. 
They are a class of ML algorithms used in decision-making problems where one has to choose between multiple options, also known as arms, and attempt to minimize a loss over time. These algorithms are able to find a good balance between exploring arms with uncertain loss and exploiting arms that have been shown to yield low losses in the past. 
This same trade-off is also important in the domain of recommender systems, where there a similar tension exists between recommending items that were popular in the past and newer, less well known options to improve diversity and avoid popularity bias \cite{jeunen:2021,jeunen:2022, 10.1145/3460231.3474618}. Bandit algorithms allow recommender systems to explore a wider range of items while exploiting items that have been previously successful. Additionally, bandit algorithms can adapt to changing user preferences and provide personalized recommendations based on a user's previous interactions with the system.
For these reasons, in the past decade, bandits have been successfully adopted in the recommender systems domain~\cite{li2010contextual}. 

Of the many variants of the multi-arm bandit problems, the ones that
are most relevant to candidate generation are the sleeping bandit
\cite{kanade2009sleeping,kleinberg2010regret} and multi-arm bandit
with multiple play problems \cite{kale2010non,uchiya2010algorithms}.
The sleeping bandit problem refers to the case when some of the arms
(candidates) are not available at some time steps, while for the
multi-arm bandit with multiple plays, more than one arm is selected at
each time step. This scenario is encountered in many commercial
applications where items are frequently out-of-stock or otherwise
unavailable. In e-commerce, many prominent companies such as Zalando,
Expedia, Airbnb, and Amazon are severely affected by this problem, yet
it does not appear to have received adequate attention in the
recommender systems literature.  Depending on the characteristics of
the specific loss function used, both sleeping bandit and multi-arm
bandits with multiple plays could be categorized as either stochastic
loss based
\cite{chatterjee2017analysis,komiyama2015optimal,chen2013combinatorial}
or adversarial loss-based bandit problems
\cite{kanade2009sleeping,kale2010non,uchiya2010algorithms,saha2020improved,neu2014online}. For
the stochastic loss, it is assumed to follow some fixed random
distributions for each arm, while no assumption has been made for the
adversarial loss except that it is usually assumed to be bounded.
Furthermore, for the sleeping bandit problem, the assumption on arm
availability can be different as well. Some prior works assume that
the set of available arms at each time step follows some unknown
stochastic distribution \cite{neu2014online,saha2020improved} while
other works assume that the arms' availability varies adversarially
\cite{kale2016hardness,kanade2014learning}.  However, as pointed out
by \cite{kleinberg2010regret,kanade2014learning,kale2016hardness}, the
problem is NP-hard when both the loss and the arm availability are
adversarial.

The discussion above points to an apparent gap in the literature
which we hope to address. The method proposed in this paper deals with
the case of sleeping bandits with multiple plays, which directly
targets the challenge of online candidate generation for
recommendation systems. Our contributions are as follows: 1) we focus
on the challenging case where the loss is adversarial but bounded and
arm availability is assumed to follow some unknown
i.i.d. distribution. 2) We propose an efficient algorithm that extends
the sleeping bandit algorithm in \cite{saha2020improved} when only one
arm is selected at each time step, and 3) we show that the theoretical
performance for the proposed algorithm is guaranteed with the regret
being upper bounded by $\bigO(kN^2\sqrt{T\log T})$, in which $k$ is
the number of arms to be selected at each time step, $N$ is the total
number of possible arms, and $T$ is the time horizon.

\section{Motivation and Problem Setup}

In the candidate generation, the candidates could be viewed as arms. And instead of choosing only one arm in the classic multi-arm bandit problem setup, we would like to select a bunch of them (with a fixed number) in order to be sorted by the ranking model afterwards. Also, for our recommendation problem, some of the candidates may not be available from time to time. This is often the case for e-commerce applications when certain items are out of stock. 
In sum, the candidate generation problem we consider is selecting a fixed number of items from the candidate pool when some candidates are unavailable from time to time.
And such an online candidate generation problem could be generalized into some variant of the multi-arm bandit problems being discussed in the last section. 
When certain arms are intermittently unavailable, it is known as the sleeping bandit problem.
In situations where multiple arms are chosen simultaneously, the multi-arm bandit problem is referred to as a multi-arm bandit with multiple plays.
Although there have been some works on either multi-arm bandit with multiple plays or sleeping multi-arm bandit, to the best of our knowledge, no prior work has dealt with the multi-arm bandit problem with both sleeping bandit and multiple play setups. Therefore, the problem setup considered in this paper is not only practical to the e-commerce recommendation problem but also new to the recommendation community.



\textbf{Problem setup:} Assume there are $N$ distinct fixed number of candidates in the candidate pool with $N$ known. At each time step $t = \{1, 2, \dots, T\}$, we are given a set of available candidates to choose from, which is represented as $S_t\subseteq [N]$. Upon receiving the available set $S_t$, we are required to select a fixed number of $k$ candidates $\Omega_t\subseteq S_t$ and then get the loss $\ell_t(i)\in [0,1]$ for each $i\in \Omega_t$. Here $\mathbf{\ell_t}\in [0,1]^k$ is the loss vector at time step $t$, and only the loss associated with the chosen candidates are revealed at each time step $t$. 
The goal of the proposed algorithm is to minimize the cumulative loss $\sum\limits_{t=1}^T \sum\limits_{i\in\Omega_t}\ell_t(i)$.

\textbf{Assumptions:}
There is no assumption on how the loss $\mathbf{\ell_t}$ is generated besides that the loss obtained at each time step $t$ is chosen obliviously and is independent of the available set $S_t$. Another assumption is for the candidate availability at each time step. We assume that each candidate's availability is independent of each other and follows the Bernoulli distribution $\mathbf{1}(i\in S_t) \sim Ber(a_i),~\forall i\in[N]$\footnote{We use $[n]$ to denote the set $\{1,2,\dots,n\}$.}. 

We measure the performance of the proposed method by the \textit{regret}, which is defined as:
\begin{equation}
\label{eq:regret_def_1}
 R_T = \max\limits_{\pi:2^{[N]}\mapsto [N]^k} \mathbb{E}\Big[\sum\limits_{t=1}^T \sum\limits_{i\in\Omega_t}\ell_t(i) - \sum\limits_{t=1}^T \sum\limits_{i\in\pi(S_t)}\ell_t(i)\Big]
\end{equation}
where $\pi:2^{[N]}\mapsto [N]^k$ is defined as the policy mapping from a set of available candidates to $k$ chosen ones and the expectation is taken w.r.t. the randomness from both the candidate availability and the algorithm. 

Let's examine the above regret definition. The chosen benchmark is the cumulative loss $\min\limits_{\pi:2^{[N]}\mapsto [N]^k}\sum\limits_{t=1}^T \sum\limits_{i\in\pi(S_t)}\ell_t(i)$, which is the minimum loss in hindsight if we know the loss for each step and the chosen candidates are fixed given the available set.
The goal of the proposed algorithm is to upper bound the above regret in a sub-linear manner w.r.t. the time horizon $T$ s.t. $\lim_{T\to\infty}\frac{R_T}{T} = \frac{o(T)}{T} = 0$, which indicates that the proposed algorithm is as good as the best static policy in hindsight on average. Such regret definition is used in lots of bandit related algorithms such as \cite{auer2000using,auer2002finite,kanade2009sleeping,xia2016budgeted,uchiya2010algorithms,komiyama2015optimal} to name a few.

In the next sections, we will show and discuss the algorithm designed to solve the above adversarial sleeping bandit with multiple plays problem and prove its sub-linear theoretical regret guarantee.

\section{Algorithm}

\begin{algorithm}[tb]
    \caption{Sleeping EXP3 for bandit problems with multiple plays}
    \label{alg:sleep_exp3_mp}
\begin{algorithmic}[1]
    \STATE {\bfseries Input:} $k$: number of bandits to select, $\lambda_t$: scale parameter, $N$: total number of bandits, $\delta > 0$: confidence parameter.
    \STATE {\bfseries Initialization:} $w_1(i) = 1, \text{for}~ i = 1, 2, \dots, N$.
    \FOR{$t=1$ {\bfseries to} $T$}
    \STATE Define $\mathbf{q}_t^S$ as the scaled probability vector after projection, which is the output of Algorithm \ref{alg:cap_prob} given the input of $N$, $k$, available set $S$, and weight vector $\mathbf{w_t}$.
    
    \STATE Receive the available bandit set $S_t\in [N]$.

    \STATE Obtain the scaled probability vector $\mathbf{q}_t^{S_t}$ from the Algorithm \ref{alg:cap_prob} given the input of $N$, $k$, available set $S_t$, and weight vector $\mathbf{w}_t$.
    \STATE Decompose the scaled probability vector $\mathbf{q}_t^{S_t}$ as the convex combination $\mathbf{q}_t^{S_t} = \sum_{\Omega} q_{\Omega}\mathbf{1}_{\Omega}$ by calling Algorithm \ref{alg:mix_decompose} given the input of $k$ and $\mathbf{q}_t^{S_t}$.

    \STATE Randomly choose the bandit arms set $\Omega_t$ with probability $q_{\Omega_t}$.
    
    \STATE Play the selected bandit arms in $\Omega_t$ and receive the loss $\ell_t(i)\in [0,1]$, for $i \in \Omega_t$.

    \STATE Compute the estimated joint probability for each arm $i\in[N]$ as:
        \begin{subequations}
        \label{eq:joint_estimation}
            \begin{align}
            \hat{a}_t(i) &= \frac{\sum_{j=1}^t\mathbf{1}(i\in S_j)}{t}\\
            P_{\hat{a}}(S) &= \Pi_{i=1}^N \hat{a}_t(i)^{\mathbf{1}(i\in S)} (1-\hat{a}_t(i))^{1-\mathbf{1}(i\in S)}\\
            \hat{q}_t(i) &= \sum_{S\in 2^{[N]}}P_{\hat{a}}(S)q_t^S(i)\label{eq:joint_prob_original}
            \end{align}
        \end{subequations}

    \STATE Compute the estimated loss as:
        \begin{equation}
        \label{eq:loss_estimate}
            \hat{\ell}_t(i) = \begin{cases}
                           \frac{\ell_t(i)}{\hat{q}_t(i)+\lambda_t}, & \text{if $i\in \Omega_t$} \\
                           0, & \text{otherwise}
                           \end{cases}
        \end{equation}

    \STATE Update the weight vector as:
    \begin{equation}
        w_{t+1}(i) = w_t(i)\exp(-\eta\hat{\ell}_t(i)), \forall i\in[N]
    \end{equation}
        
    \ENDFOR
\end{algorithmic}
\end{algorithm}

\begin{algorithm}[tb]
    \caption{Capped probability calculation}
    \label{alg:cap_prob}
\begin{algorithmic}[1]
     \STATE {\bfseries Input:} number of possible bandits $N$, number of bandits to select $k$, available set $S$, weight vector $\mathbf{w}$.
     \STATE Calculate the probability vector $\mathbf{p}$ as:
        \begin{equation}
        \nonumber
            p(i) = \begin{cases}
                           \frac{w(i)}{\sum_{j\in S}w(j)}, & \text{if $i\in S$} \\
                           0, & \text{otherwise}
                           \end{cases}
        \end{equation}
     
     \IF{$\argmax_{i\in S}p(i)\ge \frac{1}{k}$}
            \STATE Do the capping projection below to have $\mathbf{p} \in\mathcal{P}_k$ (Algorithm 4 in \cite{warmuth2008randomized}).
            \STATE Let $\mathbf{p}^{\downarrow}$ index the vector $\mathbf{p}$ in decreasing order.
            \STATE i = 1
            \REPEAT
            \STATE $\hat{\mathbf{p}} = \mathbf{p}$
            \STATE $\hat{\mathbf{p}}^{\downarrow}(j) = \frac{1}{k}$, for $j=1,\dots, i$
            \STATE $\hat{\mathbf{p}}^{\downarrow}(j) = \frac{k-i}{k}\frac{\hat{\mathbf{p}}^{\downarrow}(j)}{\sum_{l=i+1}^N\hat{\mathbf{p}}^{\downarrow}(l)}$, for $j=i+1,\dots,N$
            \STATE $i = i + 1$
            \UNTIL{$\max \hat{\mathbf{p}}\le \frac{1}{k}$}
    \ELSE
        \STATE $\hat{\mathbf{p}} = \mathbf{p}$
    \ENDIF
    \FOR{$i \in [N]$}
    \IF{$i \not\in S$}
    \STATE $q^S(i) = 0$
    \ELSE
    \STATE 
    $q^S(i) = k\hat{p}(i)$
    \ENDIF
    \ENDFOR
    \RETURN scaled probability vector $\mathbf{q^S}$
\end{algorithmic}
\end{algorithm}

\begin{algorithm}[tb]
    \caption{Mixture decomposition adapted from \cite{warmuth2008randomized}}
    \label{alg:mix_decompose}
\begin{algorithmic}[1]
     \STATE {\bfseries Input:} $1\le k\le N$ and scaled probability vector $\mathbf{p}$ satisfying $\mathbf{p}/k \in \mathcal{P}_k$.
     \WHILE{there is an $i$ with $p(i)>0$}
     \STATE Let $\mathbf{r}$ be a \{0,1\}-corner for a subset of $k$ non-zero components of $\mathbf{p}$ that includes all components of $p(i) = 1$.
     \STATE Let $s$ be the smallest of the $k$ chosen components of $\mathbf{r}$ in $\mathbf{p}$ and $l$ be the largest value of the remaining $N-k$ components in $\mathbf{p}$.
     \STATE Set $\mathbf{p} = \mathbf{p} - \underbrace{\min(s,1-l)}_{q}\mathbf{r}$ and \textbf{output} $q\mathbf{r}$ to be part of the decomposition. 
     \ENDWHILE
\end{algorithmic}
\end{algorithm}

Our proposed algorithm is shown in Algorithm \ref{alg:sleep_exp3_mp}.
Although the adversarial sleeping bandits with multiple plays problem considered in this paper is new to the best of our knowledge, there have been prior works tackling either adversarial bandits with multiple plays \cite{uchiya2010algorithms,kale2010non} or adversarial sleeping bandits \cite{saha2020improved,kleinberg2010regret} as mentioned in the Introduction.

The proposed Algorithm \ref{alg:sleep_exp3_mp} follows the idea from \cite{warmuth2008randomized} to deal with the multiple plays setting. In the classic bandit setting, only one arm is selected at each time step and it competes with the best fixed one arm in hindsight. The classic Multiplicative Weight algorithm \cite{herbster2001tracking} (and EXP3 for bandit case) is guaranteed to have $O(\sqrt{T})$ upper bound for the regret $\mathbb{E}[\sum_{t=1}^T \ell_t(i_t) - \sum_{t=1}^T \ell_t(j)], ~\forall j$. This could be reformulated as $\sum_{t=1}^T \mathbf{\ell}_t^{\top} \mathbf{p}_t - \sum_{t=1}^T \mathbf{\ell}_t^{\top} \mathbf{p}$, where both $\mathbf{p}_t$ and $\mathbf{p}$ belong to the probability simplex $\mathcal{P} = \{\mathbf{p}: \sum_i p(i) = 1 ~\text{and} ~0\le p(i)\le 1\}$. And $\mathbf{p}$ represents the corner cases where $p(j) = 1$ and $p(i) = 0, \forall i\neq j$. 

The idea from \cite{warmuth2008randomized} said that we could still have $O(\sqrt{T})$ bound if we change the domain from the probability simplex to the capped case $\mathcal{P}_k\subseteq\mathcal{P}$, which is defined as $\mathcal{P}_k = \{\mathbf{q}: \sum_i q(i) = 1 ~\text{and} ~0\le q(i)\le 1/k\}$. And accordingly, the $\mathbf{p}_t$ and $\mathbf{p}$ should belong to $\mathcal{P}_k$, and in particular, $\mathbf{p}$ would be the corner cases where only $k$ non-zero components in $\mathbf{p}$ being equal to $1/k$. Then the regret could be written as $\sum_{t=1}^T \mathbf{\ell}_t^{\top} \mathbf{p}_t - \sum_{t=1}^T \mathbf{\ell}_t^{\top} \mathbf{p}$ = $\sum_{t=1}^T \mathbf{\ell}_t^{\top} \mathbf{p}_t -\sum_{t=1}^T \sum_{j\in\Omega}\ell_t(j) \frac{1}{k}$, where $\mathbf{p}_t \in \mathcal{P}_k$ and $\Omega$ represents the arbitrary subset with $k$ components. If you compare this regret formula with our regret definition in Eq.~(\ref{eq:regret_def}), the second term in Eq.~(\ref{eq:regret_def}) is scaled by $k$. To make the first term have similar structure, \cite{warmuth2008randomized} proved that any vector $\mathbf{p}\in\mathcal{P}_k$ could be decomposed as a convex combination $\mathbf{p} = \sum_S w_S\mathbf{1}_S\frac{1}{k}$, where $S$ represents the set with $k$ components and $0\le w_S\le 1, ~\sum w_S = 1$. As a result, the first term could be reformulated as $\sum_{t=1}^T \mathbf{\ell}_t^{\top} \mathbf{p}_t = \sum_{t=1}^T \sum_S\mathbf{\ell}_tw_S\mathbf{1}_S\frac{1}{k} = \mathbb{E}[\sum_{t=1}^T\sum_{i\in\Omega_t}\ell_t(i)\frac{1}{k}]$, where the RHS is obtained by randomly choosing the $k$ components subset $\Omega_t$ with the probability $w_{\Omega_t}$. Then the regret formula could be rewritten as $\mathbb{E}[\sum_{t=1}^T\sum_{i\in\Omega_t}\ell_t(i)\frac{1}{k}-\sum_{t=1}^T \sum_{j\in\Omega}\ell_t(j) \frac{1}{k}]$. And the difference between our regret in Eq.~(\ref{eq:regret_def}) and this formula is that it is scaled by $k$ and with extra expectation over the arm availability. 

To see how the above idea is leveraged in Algorithm \ref{alg:sleep_exp3_mp}, let's first see its Step 4, whose output is from Algorithm \ref{alg:cap_prob}. In this step, it first checks if the current probability vector $\mathbf{p}\in \mathcal{P}_k$. If not, it will do the projection to the capped probability vector space $\mathcal{P}_k$ mentioned above. Then the capped probability vector will be scaled by $k$, the number of bandits to select, to make sure the regret formula is the same as our definition in Eq.~(\ref{eq:regret_def}). Then for Step 7 in Algorithm \ref{alg:sleep_exp3_mp}, it does the same operation as mentioned above to decompose the scaled probability vector as a convex combination of the corners. And each corner only has k non-zero components being equal to 1 as shown in Step 3 from Algorithm \ref{alg:mix_decompose}. Then for Step 8 of Algorithm \ref{alg:sleep_exp3_mp}, the probability of choosing the arm $i$ is equal to $\mathbf{q}_t(i)$. And Step 12 of Algorithm \ref{alg:sleep_exp3_mp} follows the same step in \cite{warmuth2008randomized} to update the weights associated with the chosen arms, which is very common in the classic multiplicative weight based methods.

The other main part of the Algorithm \ref{alg:sleep_exp3_mp} is to estimate the probability of choosing one specific arm $i$ and connect it to the received loss, which is inspired by \cite{saha2020improved} with the same sleeping bandit part setup. Unlike the classic non-sleeping bandit problem, whose probability of being chosen purely depends on the algorithm itself, which is the value of $\mathbf{q}_t(i)$, the sleeping bandit problem considered in this paper also has the randomness from the arms' availability. As a result, what we do in Algorithm \ref{alg:sleep_exp3_mp} is to estimate the probabilities of the algorithm randomness and arm availability randomness jointly as shown in Eq.~(\ref{eq:joint_estimation}). 
Then we estimate the loss by Eq.~(\ref{eq:loss_estimate}), which follows the idea of the classic bandit algorithm EXP3 \cite{auer2002nonstochastic} except that there is an extra term $\lambda_t$ to try to reduce the variance of the estimation.

\textbf{Time complexity of the Algorithm \ref{alg:sleep_exp3_mp}:} Computing the probability projection in Step 4 takes $(\bigO(K\log(K)))$ time since it requires to sort the probability vector. Implementing Step 7 to decompose the scaled probability vector as a convex combination of corners requires $\bigO(K^2)$, since the decomposition has at most $K$ terms and it needs $\bigO(K)$ time for each loop \cite{warmuth2008randomized}. Step 10 is the most time-consuming step since we need to estimate the joint probability by going through all possible combinations, which takes $\bigO(2^K)$ time. As a result, the time complexity of the Algorithm \ref{alg:sleep_exp3_mp} is $\bigO(2^{k})$ per round.

\textbf{A more time efficient alternative of Step 10 in Algorithm \ref{alg:sleep_exp3_mp}:}
$\bigO(2^{k})$ per round time complexity is undesirable. Luckily, the work in \cite{saha2020improved} also proposed a more time efficient estimation of the joint probability, which could also be used in our algorithm. 
According to \cite{saha2020improved}, $\hat{q}_t(i)$ could be approximated by an empirical estimate shown as below:
\begin{equation}
\label{eq:efficient_joint}
    \tilde{q}_t(i) = \frac{1}{t}\sum_{\tau=1}^t q_t^{S_t^{(\tau)}}(i)
\end{equation}
where $S_t^{(1)},\dots,S_t^{(t)}$ are drawn independently from the empirical availability distribution $P_{\hat{a}}(S)$ at time step $t$. And $\tilde{q}_t(i)$ is an unbiased estimate of $\hat{q}_t(i)$ given the availability distribution estimate at time step $t$. Then we could use the concentration inequality to show that this unbiased estimate is very close to the true value with high probability. And for the time step $t$, the time complexity of such estimate is $\bigO(tK)$, which is much more efficient than the original one's.   

In the next section, we will show that the proposed Algorithm \ref{alg:sleep_exp3_mp} has the theoretical guarantee that could upper bound the regret defined in Eq.~(\ref{eq:regret_def}) by $\bigO(\sqrt{T\log T})$ for both the inefficient (using original Step $7$) and the efficient (by Eq.~(\ref{eq:efficient_joint})) joint probability estimation.

\section{Theoretical Results}


\begin{theorem}
\label{thm:main_thm}
If we set $\lambda_t = \min\{1,2kN\sqrt{\frac{2\log(N/\delta)}{t}}+\frac{8kN\log(N/\delta)}{3t}\}$, $\delta = N/T^2$, $\eta =  \sqrt{\ln(N/k)/NT}$, and follow the Algorithm \ref{alg:sleep_exp3_mp}, then the regret defined in Eq.~(\ref{eq:regret_def_1}) could be upper bounded as
\begin{equation}
\nonumber
\begin{array}{ll}
R_T &= \max\limits_{\pi:2^{[N]}\mapsto [N]^k} \mathbb{E}\Big[\sum\limits_{t=1}^T \sum\limits_{i\in\Omega_t}\ell_t(i) - \sum\limits_{t=1}^T \sum\limits_{i\in\pi(S_t)}\ell_t(i)\Big] 
    \le \bigO(kN^2\sqrt{T\log T})
\end{array}
\end{equation}
\end{theorem}

\begin{proof}
Let's recall first the regret definition in Eq.~(\ref{eq:regret_def_1}) as below
\begin{equation}
\nonumber
\label{eq:regret_def}
 R_T = \max\limits_{\pi:2^{[N]}\mapsto [N]^k} \mathbb{E}\Big[\sum\limits_{t=1}^T \sum\limits_{i\in\Omega_t}\ell_t(i) - \sum\limits_{t=1}^T \sum\limits_{i\in\pi(S_t)}\ell_t(i)\Big]
\end{equation}
where $\pi:2^{[N]}\mapsto [N]^k$ is defined as the policy mapping from a set of available candidates to $k$ chosen ones and the expectation is taken w.r.t. the randomness from both the candidate availability and the algorithm. 

Denote the best policy as $\pi^\star = \argmin_{\pi:2^{[N]}\mapsto [N]^k}\sum\limits_{t=1}^T \mathbb{E}_{S_t\sim P_a}\Big[\sum_{i\in\pi(S_t)}\ell_t(i)\Big]$. Then the above regret can be written as 
\begin{equation}
\label{eq:regret_reform}
\begin{array}{ll}
    R_T &= \mathbb{E}\Big[\sum\limits_{t=1}^T \sum\limits_{i\in\Omega_t}\ell_t(i) - \sum\limits_{t=1}^T \sum\limits_{i\in\pi^\star(S_t)}\ell_t(i)\Big] 
\end{array}
\end{equation}

First, let's see how to relate the term of $\mathbb{E}[\sum_{i\in\Omega_t}\ell_t(i)]$ with the term $\mathbb{E}[\hat{\ell}_t^\top \mathbf{q}_t^S]$. Since $\Omega_t$ is obtained through Step 8 in Algorithm \ref{alg:sleep_exp3_mp} by following the probability $\mathbf{q}_t^{S_t}$, $\mathbb{E}[\sum_{i\in\Omega_t}\ell_t(i)]$ is equal to $\mathbb{E}[\ell_t^\top \mathbf{q}_t^{S_t}]$. As a result, to get the relationship between $\mathbb{E}[\ell_t^\top \mathbf{q}_t^{S_t}]$ and $\mathbb{E}[\hat{\ell}_t^\top \mathbf{q}_t^S]$, we plug in the $\hat{\ell}_t$ definition along with the $k$ scaled concentration upper bound for $|q_t^\star(i) - \hat{q}_t(i)$ from Lemma 1 of \cite{saha2020improved}. 
Then the relation below can be derived by following Lemma 3 of \cite{saha2020improved}:
\begin{equation}
    \label{eq:R_T_1st_bound}
    \mathbb{E}\Big[\sum_{i\in\Omega_t}\ell_t(i)\Big] \le \mathbb{E}\Big[\hat{\ell}_t^\top \mathbf{q}_t^S\Big] + 2N\lambda_t+\frac{\delta}{\lambda_t}
\end{equation}
where $\delta \in (0,1)$ and $\lambda_t = \min\{1,2kN\sqrt{\frac{2\log(N/\delta)}{t}}+\frac{8kN\log(N/\delta)}{3t}\}$.

Next, we would like to get the relation between $\mathbb{E}[\hat{\ell}_t(i)]$ and $\ell_t(i)$ for any $i\in[N]$. This relation has been created by Lemma 4 in \cite{saha2020improved} and we just rewrite it to fit into our problem:
\begin{lemma}
    Let $\delta \in (0,1)$. Define $\hat{\ell}_t$ as in Eq.~(\ref{eq:loss_estimate}) and assume that $i\in\Omega_t$ is drawn according to $\mathbf{q}_t^{S_t}$ as generated in Algorithm \ref{alg:sleep_exp3_mp}. Then for any $i\in[N]$,
    \begin{equation}
    \label{eq:R_T_2nd_bound}
    \mathbb{E}[\hat{\ell}_t(i)] \le \ell_t(i) + \frac{\delta}{\lambda_t}
    \end{equation}
    where 
    $\lambda_t = \min\{1,2kN\sqrt{\frac{2\log(N/\delta)}{t}}+\frac{8kN\log(N/\delta)}{3t}\}$.
\end{lemma}

Then regret in Eq.~(\ref{eq:regret_reform}) could be rewritten as below after plugging in the relations from Eq.~(\ref{eq:R_T_1st_bound}) and Eq.~(\ref{eq:R_T_2nd_bound})
\begin{equation}
\begin{array}{ll}
R_T &\le \sum\limits_{t=1}^T\Bigg[\mathbb{E}\Big[\hat{\ell}_t^\top \mathbf{q}_t^S - \hat{\ell}_t^\top \mathbf{1}_{\pi^\star(S_t)}\Big] + 2N\lambda_t+\frac{\delta}{\lambda_t} + \frac{k\delta}{\lambda_t}\Bigg] \\
&= \sum\limits_{t=1}^T\mathbb{E}\Big[\hat{\ell}_t^\top \mathbf{q}_t^S - \hat{\ell}_t^\top \mathbf{1}_{\pi^\star(S)}\Big] + 2N\sum\limits_{t=1}^T\lambda_t + (k+1)\delta \sum\limits_{t=1}^T\frac{1}{\lambda_t} \\
&= \mathbb{E}\Bigg[\sum\limits_{t=1}^T\mathbb{E}\Big[\hat{\ell}_t^\top \mathbf{q}_t^S - \hat{\ell}_t^\top \mathbf{1}_{\pi^\star(S)}\Big|S\Big]\Bigg] +2N\sum\limits_{t=1}^T\lambda_t + (k+1)\delta \sum\limits_{t=1}^T\frac{1}{\lambda_t}
\end{array}
\end{equation}
where we use the assumption that the bandits' availability is independent of the generated loss.

As a result, we could upper bound $\sum\limits_{t=1}^T\mathbb{E}\Big[\hat{\ell}_t^\top \mathbf{q}_t^S - \hat{\ell}_t^\top \mathbf{1}_{\pi^\star(S)}\Big|S\Big]$ assuming that the available set $S$ is fixed. And as explained in the Algorithm section, since $\mathbf{q}_t^S/k\in \mathcal{P}_k$ and $\mathbf{1}_{\pi^\star(S)}/k\in\mathcal{P}_k$, where $\mathcal{P}_k = \{\mathbf{q}: \sum_i q(i) = 1 ~\text{and} ~0\le q(i)\le 1/k\}$ is a probability vector space, we could apply the Theorem 3.1 of \cite{kale2010non} to upper bound our $k$-scaled version of the original form, which can be formulated as our case below:
\begin{equation}
\nonumber
    \sum\limits_{t=1}^T\mathbb{E}\Big[\hat{\ell}_t^\top \mathbf{q}_t^S - \hat{\ell}_t^\top \mathbf{1}_{\pi^\star(S)}\Big|S\Big] \le \frac{k\ln(N/k)}{\eta} + k\eta \sum\limits_{t=1}^T\big(\hat{\ell}_t^2\big)^\top \mathbf{q}_t^S
\end{equation}

Then the regret could be upper bounded as below
\begin{equation}
    R_T \le k\eta \sum\limits_{t=1}^T\mathbb{E}\Big[\big(\hat{\ell}_t^2\big)^\top \mathbf{q}_t^S\Big] + \frac{k\ln(N/k)}{\eta}+2N\sum\limits_{t=1}^T\lambda_t + (k+1)\delta \sum\limits_{t=1}^T\frac{1}{\lambda_t}
\end{equation}

Lemma 5 in \cite{saha2020improved} helps further upper bound the term $\mathbb{E}\Big[\big(\hat{\ell}_t^2\big)^\top \mathbf{q}_t^S\Big]$ by $N+\frac{\delta}{\lambda_t^2}$ for the same $\lambda_t$ as before, which results in the following inequality:
\begin{equation}
\label{eq:proof_ineq}
\begin{array}{ll}
    R_T &\le kNT\eta + k\eta\delta \sum\limits_{t=1}^T\frac{1}{\lambda_t^2} +  \frac{k\ln(N/k)}{\eta}+2N\sum\limits_{t=1}^T\lambda_t + (k+1)\delta \sum\limits_{t=1}^T\frac{1}{\lambda_t} \\
       &\le kNT\eta + \frac{k\ln(N/k)}{\eta} + 2N\sum\limits_{t=1}^T\lambda_t + (2k+1)\delta \sum\limits_{t=1}^T\frac{1}{\lambda_t^2}
\end{array}
\end{equation}
where $\lambda_t = \min\{1,2kN\sqrt{\frac{2\log(N/\delta)}{t}}+\frac{8kN\log(N/\delta)}{3t}\}$, $\eta \le 1$, and the last inequality follows.   

Using $\delta = N/T^2$, we could see that it takes $\bigO(\log(T))$ time steps to have $\lambda_t = 2kN\sqrt{\frac{2\log(N/\delta)}{t}}+\frac{8kN\log(N/\delta)}{3t}$. As a result, we could just focus on this ongoing time period, given the regret from $t=1$ to $t = \bigO(\log(T))$ could be upper bounded by $\bigO(\log(T))$.

For the first two terms on RHS, by using $\eta = \min\{1, \sqrt{\ln(N/k)/NT}\}$, it can be upper bounded as $2k\sqrt{NT\ln(N/k)}$.

For the third term, with $\lambda_t = 2kN\sqrt{\frac{2\log(N/\delta)}{t}}+\frac{8kN\log(N/\delta)}{3t}$, $\sum_{t=1}^T \lambda_t \le 4kN\sqrt{T\log T} + 16kN\log T/3(1+\log T)$.

For the last term $\sum\limits_{t=1}^T\frac{1}{\lambda_t^2}$, since $\lambda_t \ge 2kN\sqrt{\frac{2\log(N/\delta)}{t}}$, $\sum\limits_{t=1}^T\frac{1}{\lambda_t^2} \le \frac{T^2}{8k^2N^2}$.

After putting the above inequalities together, we have $R_T \le \bigO(kN^2\sqrt{T\log T})$.
\end{proof}


\begin{lemma}
If we use the alternative more efficient estimation in Eq.~(\ref{eq:efficient_joint}) to replace the update of Eq.~(\ref{eq:joint_estimation}) in Algorithm \ref{alg:sleep_exp3_mp}, the regret can be bounded below after setting $\lambda_t = \min\big\{1,4kN\sqrt{\frac{\log(2N/\delta)}{t}}+\frac{8kN\log(2N/\delta)}{3t}\big\}$, $\delta = N/T^2$, $\eta =  \sqrt{\ln(N/k)/NT}$:
\begin{equation}
\nonumber
\begin{array}{ll}
R_T &= \max\limits_{\pi:2^{[N]}\mapsto [N]^k} \mathbb{E}\Big[\sum\limits_{t=1}^T \sum\limits_{i\in\Omega_t}\ell_t(i) - \sum\limits_{t=1}^T \sum\limits_{i\in\pi(S_t)}\ell_t(i)\Big] 
    \le \bigO(kN^2\sqrt{T\log T})
\end{array}
\end{equation}

\end{lemma}

\begin{proof}[Proof sketch]    
Most of the proof follows from Theorem \ref{thm:main_thm} except that
we now need to show that the alternative estimate of the joint
probability still concentrates around the true probability with a very
similar bound. Since we are using the same idea as in
\cite{saha2020improved}, this concentration bound has been proved in
Lemma 6 therein. Accordingly, we need to change the $\lambda_t$ value
to $\lambda_t =
\min\big\{1,4kN\sqrt{\frac{\log(2N/\delta)}{t}}+\frac{8kN\log(2N/\delta)}{3t}\big\}$. This
results in the same inequality as in Eq.~(\ref{eq:proof_ineq}) along
with similar bounds for each term. Hence, the regret is upper bounded
as above.
\end{proof}

\section{Conclusion and Future Work}

This paper proposes a novel multi-arm bandit algorithm for solving the
online candidate generation problem. We prove theoretical performance
guarantees for the algorithm, specifically that regret is upper
bounded by $\bigO(kN^2\sqrt{T\log T})$, where $k$, $N$, and $T$
represent the number of arms selected per time step, the total number
of arms, and the time horizon, respectively.  Compared to the classic
multi-arm bandit problem, online candidate generation is similar in
terms of online information feedback, but retains the unique features
of arms' unavailability and multiple arm selection. The proposed
algorithm extends the existing sleeping bandit algorithm, which can
only handle arms' unavailability, to the multiple-play setting, where
multiple arms are selected at each time step.  At this early stage,
this work focuses solely on the theoretical understanding of the
proposed algorithm adapted to online candidate generation.  In the
next stage, we will perform offline evaluation of the approach both on
public and proprietary datasets, followed by an A/B test to measure
online performance in our production environment.


\section*{Acknowledgments}
The authors would like to thank the reviewers for their valuable feedback. We would also want to thank the team leads from Expedia Group: Yi Cao, Yonghai Li, Albert Nedvall, and Zoe Yang for their generous support on this work.

\bibliographystyle{unsrt}
\bibliography{ref, ref_ludo}

\begin{thebibliography}{10}

\bibitem{Ricci2022}
Francesco Ricci, Lior Rokach, and Bracha Shapira.
\newblock {\em Recommender Systems: Techniques, Applications, and Challenges},
  pages 1--35.
\newblock Springer US, New York, NY, 2022.

\bibitem{Smith2017}
Brent Smith and Greg Linden.
\newblock Two decades of recommender systems at amazon.com.
\newblock {\em IEEE Internet Computing}, 2017.

\bibitem{amatriain2015recommender}
Xavier Amatriain and Justin Basilico.
\newblock Recommender systems in industry: A netflix case study.
\newblock {\em Recommender systems handbook}, pages 385--419, 2015.

\bibitem{10.1145/3460231.3474618}
Andreas Gr\"{u}n and Xenija Neufeld.
\newblock Challenges experienced in public service media recommendation
  systems.
\newblock In {\em Proceedings of the 15th ACM Conference on Recommender
  Systems}, RecSys '21, page 541–544, New York, NY, USA, 2021. Association
  for Computing Machinery.

\bibitem{covington2016deep}
Paul Covington, Jay Adams, and Emre Sargin.
\newblock Deep neural networks for youtube recommendations.
\newblock In {\em Proceedings of the 10th ACM conference on recommender
  systems}, pages 191--198, 2016.

\bibitem{rendle2008online}
Steffen Rendle and Lars Schmidt-Thieme.
\newblock Online-updating regularized kernel matrix factorization models for
  large-scale recommender systems.
\newblock In {\em Proceedings of the 2008 ACM conference on Recommender
  systems}, pages 251--258, 2008.

\bibitem{auer2000using}
Peter Auer.
\newblock Using upper confidence bounds for online learning.
\newblock In {\em Proceedings 41st Annual Symposium on Foundations of Computer
  Science}, pages 270--279. IEEE, 2000.

\bibitem{auer2002finite}
Peter Auer, Nicolo Cesa-Bianchi, and Paul Fischer.
\newblock Finite-time analysis of the multiarmed bandit problem.
\newblock {\em Machine learning}, 47:235--256, 2002.

\bibitem{auer2002nonstochastic}
Peter Auer, Nicolo Cesa-Bianchi, Yoav Freund, and Robert~E Schapire.
\newblock The nonstochastic multiarmed bandit problem.
\newblock {\em SIAM journal on computing}, 32(1):48--77, 2002.

\bibitem{vermorel2005multi}
Joannes Vermorel and Mehryar Mohri.
\newblock Multi-armed bandit algorithms and empirical evaluation.
\newblock In {\em Machine Learning: ECML 2005: 16th European Conference on
  Machine Learning, Porto, Portugal, October 3-7, 2005. Proceedings 16}, pages
  437--448. Springer, 2005.

\bibitem{cesa2006prediction}
Nicolo Cesa-Bianchi and G{\'a}bor Lugosi.
\newblock {\em Prediction, learning, and games}.
\newblock Cambridge university press, 2006.

\bibitem{jeunen:2021}
Olivier Jeunen and Bart Goethals.
\newblock Pessimistic reward models for off-policy learning in recommendation.
\newblock In {\em Proceedings of the 15th ACM Conference on Recommender
  Systems}, RecSys '21, page 63–74, New York, NY, USA, 2021. Association for
  Computing Machinery.

\bibitem{jeunen:2022}
Olivier Jeunen and Bart Goethals.
\newblock Pessimistic decision-making for recommender systems.
\newblock {\em ACM Trans. Recomm. Syst.}, oct 2022.
\newblock Just Accepted.

\bibitem{li2010contextual}
Lihong Li, Wei Chu, John Langford, and Robert~E Schapire.
\newblock A contextual-bandit approach to personalized news article
  recommendation.
\newblock In {\em Proceedings of the 19th international conference on World
  wide web}, pages 661--670, 2010.

\bibitem{kanade2009sleeping}
Varun Kanade, H~Brendan McMahan, and Brent Bryan.
\newblock Sleeping experts and bandits with stochastic action availability and
  adversarial rewards.
\newblock In {\em Artificial Intelligence and Statistics}, pages 272--279.
  PMLR, 2009.

\bibitem{kleinberg2010regret}
Robert Kleinberg, Alexandru Niculescu-Mizil, and Yogeshwer Sharma.
\newblock Regret bounds for sleeping experts and bandits.
\newblock {\em Machine learning}, 80(2-3):245--272, 2010.

\bibitem{kale2010non}
Satyen Kale, Lev Reyzin, and Robert~E Schapire.
\newblock Non-stochastic bandit slate problems.
\newblock {\em Advances in Neural Information Processing Systems}, 23, 2010.

\bibitem{uchiya2010algorithms}
Taishi Uchiya, Atsuyoshi Nakamura, and Mineichi Kudo.
\newblock Algorithms for adversarial bandit problems with multiple plays.
\newblock In {\em Algorithmic Learning Theory: 21st International Conference,
  ALT 2010, Canberra, Australia, October 6-8, 2010. Proceedings 21}, pages
  375--389. Springer, 2010.

\bibitem{chatterjee2017analysis}
Aritra Chatterjee, Ganesh Ghalme, Shweta Jain, Rohit Vaish, and Y~Narahari.
\newblock Analysis of thompson sampling for stochastic sleeping bandits.
\newblock In {\em UAI}, 2017.

\bibitem{komiyama2015optimal}
Junpei Komiyama, Junya Honda, and Hiroshi Nakagawa.
\newblock Optimal regret analysis of thompson sampling in stochastic
  multi-armed bandit problem with multiple plays.
\newblock In {\em International Conference on Machine Learning}, pages
  1152--1161. PMLR, 2015.

\bibitem{chen2013combinatorial}
Wei Chen, Yajun Wang, and Yang Yuan.
\newblock Combinatorial multi-armed bandit: General framework and applications.
\newblock In {\em International conference on machine learning}, pages
  151--159. PMLR, 2013.

\bibitem{saha2020improved}
Aadirupa Saha, Pierre Gaillard, and Michal Valko.
\newblock Improved sleeping bandits with stochastic action sets and adversarial
  rewards.
\newblock In {\em International Conference on Machine Learning}, pages
  8357--8366. PMLR, 2020.

\bibitem{neu2014online}
Gergely Neu and Michal Valko.
\newblock Online combinatorial optimization with stochastic decision sets and
  adversarial losses.
\newblock {\em Advances in Neural Information Processing Systems}, 27, 2014.

\bibitem{kale2016hardness}
Satyen Kale, Chansoo Lee, and D{\'a}vid P{\'a}l.
\newblock Hardness of online sleeping combinatorial optimization problems.
\newblock {\em Advances in Neural Information Processing Systems}, 29, 2016.

\bibitem{kanade2014learning}
Varun Kanade and Thomas Steinke.
\newblock Learning hurdles for sleeping experts.
\newblock {\em ACM Transactions on Computation Theory (TOCT)}, 6(3):1--16,
  2014.

\bibitem{xia2016budgeted}
Yingce Xia, Tao Qin, Weidong Ma, Nenghai Yu, and Tie-Yan Liu.
\newblock Budgeted multi-armed bandits with multiple plays.
\newblock In {\em IJCAI}, volume~6, pages 2210--2216, 2016.

\bibitem{warmuth2008randomized}
Manfred~K Warmuth and Dima Kuzmin.
\newblock Randomized online pca algorithms with regret bounds that are
  logarithmic in the dimension.
\newblock {\em Journal of Machine Learning Research}, 9(Oct):2287--2320, 2008.

\bibitem{herbster2001tracking}
Mark Herbster and Manfred~K Warmuth.
\newblock Tracking the best linear predictor.
\newblock {\em Journal of Machine Learning Research}, 1(281-309):10--1162,
  2001.

\end{thebibliography}

\end{document}